# Quantum-Inspired Genetic Optimization for Patient Scheduling in Radiation Oncology

Akira SaiToh[1*], Arezoo Modiri[2], Amit Sawant[2], Robabeh Rahimi[2*]

[1] Department of Computer and Information Sciences, Sojo University, Kumamoto, Kumamoto, Japan

[2] Department of Radiation Oncology, University of Maryland School of Medicine, Baltimore, Maryland, United States of America

* Corresponding authors
Emails: st@cis.sojo-u.ac.jp (AST), RRahimi@som.umaryland.edu (RR)

# Abstract

Among the genetic algorithms generally used for optimization problems in the recent decades, quantum-inspired variants are known for fast and high-fitness convergence and small resource requirement. Here the application to the patient scheduling problem in proton therapy is reported. Quantum chromosomes are tailored to possess the superposed data of patient IDs and gantry statuses. Selection and repair strategies are also elaborated for reliable convergence to a clinically feasible schedule although the employed model is not complex. Clear advantage in population size is shown over the classical counterpart in our numerical results for both a medium-size test case and a large-size practical problem instance. It is, however, observed that program run time is rather long for the large-size practical case, which is due to the limitation of classical emulation and demands the forthcoming true quantum computation. Our results also revalidate the stability of the conventional classical genetic algorithm.

# 1. Introduction

Radiation therapy has been one of the primary modalities for cancer treatment [1]. Proton therapy is its specialized form distinguished by precise targeting of a tumor tissue and minimal radiation dose delivered to tissues distal to the target [2,3]. It has been of practical interest in this field to optimize the daily schedule of treatment rooms for the benefit of patients and the efficiency of equipment usage in light of the running cost of gantries [4]. This problem belongs to a class of partitioning and job scheduling problems that are believed to be computationally difficult to solve in a deterministic manner in general [5]. In the literature [6,7], Monte Carlo optimizations have been used and successful for generating improved daily schedules.

The schedule optimization problem, in general, has a long history and has been tackled in several different approaches [5,8]. One of the most successful approaches was a genetic algorithm [9-11]. A schedule is typically represented by an array of cells and each cell represents a status at the corresponding time slot. Such an array is called chromosome or individual, and a set of chromosomes is called generation or population. The generation evolves under the (simulated) natural selection, the crossover, and the mutation to eventually enlarge the population size of desirable chromosomes that have high fitness values under the environment. It is expected that the fitness improvement reaches a convergence within practical time and the chromosome with the best fitness after convergence is used as an optimized schedule.

A variant of genetic algorithms, quantum-inspired genetic algorithms [12-15], has been known for fast and high-fitness convergence and small (computational) resource requirement originated from the quantum nature albeit virtually realized without real quantum resources. They are characterized by the use of quantum states representing chromosomes and the repair operations enhancing desirable amplitudes in the states with unitary rotations. It has been discussed that migrations of chromosomes between groups are useful for keeping the data of high-fitness chromosomes [13]; then later discussed that the pair-swap strategy may eliminate such migrations and still keep the high-fitness chromosome data [14]. These extra strategies utilized the saved best chromosomes: the former strategy used the group best and the latter used the individual personal best. Hence one may need to doubly count the number of chromosomes in the latter strategy for rigorous resource evaluation. We do not employ either strategy in this study as they are non-essential for importing quantum nature.

In this contribution, we propose a quantum-inspired genetic algorithm dedicated to the patient scheduling in proton therapy. A classical genetic algorithm is first introduced in a conventional manner in which each chromosome represents the entire daily schedule using patient IDs and gantry statuses. A quantum genetic algorithm is then introduced as an extension where each quantum chromosome represents a daily schedule in which a cell for each time slot keeps a superposition of available patient IDs and that of gantry statuses. Our numerical results will show the advantage of quantum-inspired one in the required population size, while not in the program run time. The results will be discussed mainly in view of computational resources.

The remaining part of this paper is organized as follows: In section 2, the scheduling problem of our interest is introduced as a definite model. Our classical and quantum-inspired genetic algorithms based on the model are described in section 3. The numerical results using the algorithms are shown in section 4. Section 5 gives the discussion on the results. The conclusion is given in section 6.

## 2. Model of Patient Scheduling

The scope of our study is the management of daily schedule of a proton therapy center operating with common and conventional apparatus. We assume that patients are identified by their unique IDs and only one treatment session is conducted for each ID in the daily schedule. The twice-daily (Bi-daily, BID) treatment is excluded for simplicity, but this does not compromise generality as two distinct IDs may be assigned to one patient. We also assume that we have multiple gantries operating independently. Although they share a single cyclotron or synchrotron, this does not limit the performance and availability of each gantry's operations in a standard setup. The operational states, namely, statuses, of a gantry are described in Table 1, where we follow the directions in Sakae et al.'s work [7]. Durations (or time consumptions) of the statuses are also listed in Table 1. Although some radiotherapy facilities may have an imaging room for patient positioning before conveyance to a gantry, most modern clinics have in-room imaging, e.g., gantry-mounted on-board imagers, MRI-LINAC [16], etc., to ensure accurate patient localization. For this reason, we assume that there is no separate offline imaging procedure performed in another room before each operational fraction. For the same reason, we also omit advanced time-shortening options like the shuttle-based conveyance system [17].

**Table 1. Symbols and descriptions of gantry statuses during treatment for one patient.** Duration is coarse-grained with a unit time 1 minute.

| Symbol / Color | Status description | Duration [min] |
|---|---|---|
| G_IDL | Idle | 1 |
| G_R | Ready | 1 |
| G_WP | Waiting for a Patient | 3 |
| G_AT | Adjusting for Target based on position requests | 15 |
| G_WC | Waiting for a Control to be ready for irradiation | 1 |
| G_WA | Waiting for Accelerator to be ready | 1 |
| G_IR | On IRradiation | 1 |
| G_PD | Preparation for the Dispose and the next control | 4 |

The expected transition of the statuses during the patient treatment is as follows: G_IDL → G_R → G_WP → G_AT → G_WC → G_WA → G_IR → G_PD → G_IDL → ⋯ . Suppose we have $n_g$ gantries labeled as 0, 1, …, $n_g$-1. For each gantry, we have time slots 0, 1, …, $n_t$-1 and each time slot has an assigned patient ID except for the idle time. In total, the transition of statuses is depicted as Fig 1 in which statuses are represented by their colors found in Table 1.

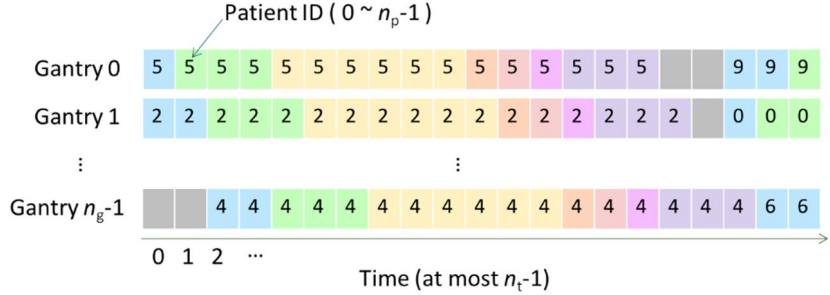

**Fig 1. Status transitions in the gantries.** Each status is presented by its color (see Table 1).

# 3. Algorithm Design

The optimization of the patient schedule depicted in Fig 1 is performed by an evolutionary way; we will first explore a standard classic genetic algorithm and second a quantum-inspired genetic algorithm with tailored quantum chromosomes.

## 3.1 Classical Genetic Algorithm

For the classical genetic algorithm, we employ the conventional ranking rule for selection and a single-point crossover, while we put the entire daily schedule in a chromosome. The flow of the algorithm is depicted in Fig 2.

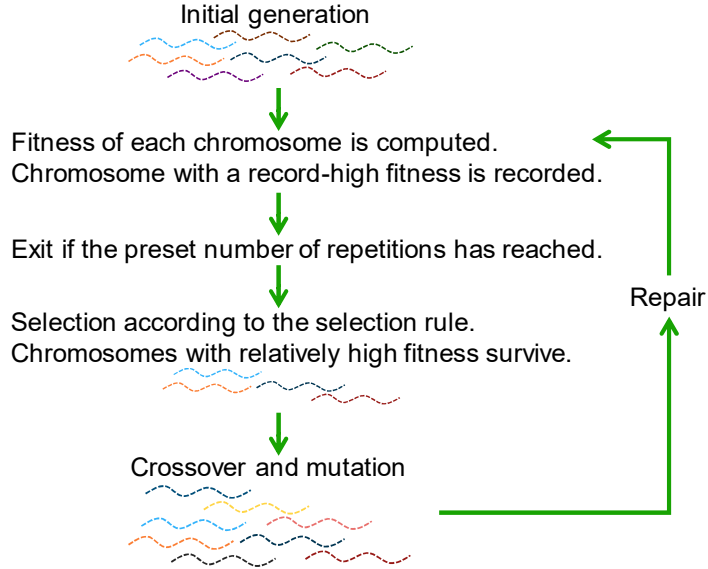

**Fig 2. The flow of the classical genetic algorithm for our schedule optimization.**

### 3.1.1 Chromosome Design

A chromosome is designed to include the entire daily schedule as depicted in Fig 3. It has $n_g$ tracks each of which represents the daily schedule for the gantry. Each track consists of at most $n_t$ time slots, each of which contains the patient ID under the therapy (vacant for idle time) and the status of the gantry.

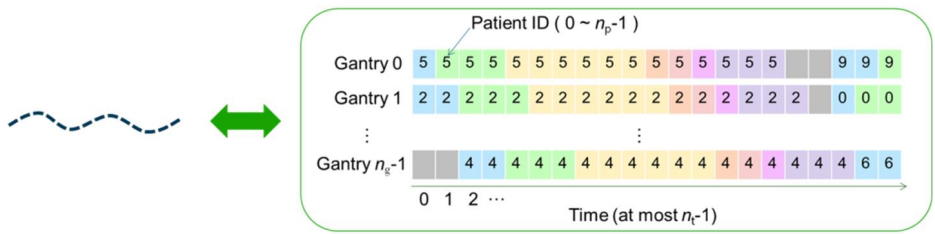

**Fig 3. The chromosome design** (see the text for details).

The initial generation consists of 10 or its multiple randomly generated chromosomes in which each cell of each track has a randomly assigned patient ID and a random status.

### 3.1.2 Selection Rule and the Fitness

At the selection, a so-called ranking strategy is employed: chromosomes are ordered along with their fitness values and the upper ones survive with the surviving ratio $r_s$. In case we have a preset maximum population size $N_{max}$, the number of surviving chromosomes is also bounded by it.

The fitness values are calculated for individual chromosomes using the penalty and benefit scores in Table 2. The fitness value is the sum of the benefit values subtracted by the sum of the penalty values.

**Table 2. List of penalty and benefit scores.** Note that every occurrence of each penalty/benefit case is counted (not just once, in general) by seeking from the head to the tail of a chromosome.

| Penalty | Score |
|---|---|
| Patient conflict (same IDs for multiple gantries) | 20 |
| Non-consecutive statuses during the expected duration (duration from G_R to the end of G_PD) | 20 |
| Multiple treatments for the same patient ID | 28 |
| Interruption of a patient by another patient | 12 |
| Time consumption penalty per time slot | 1.5 |

| Benefit | Score |
|---|---|
| Consecutive statuses as expected for their duration | 3.0 |
| Statuses are performed in the expected order | 20 |
| A therapy completed for a patient | 20 |

### 3.1.3 Crossover

The crossover operation is conducted in the following manner: We have a preset crossover ratio $r_c$ typically in the range [0.05, 0.50]. Let us assume that the generation prior to the crossover consists of $N$ chromosomes. Then, we make $r_c N/2$ pairs randomly and for each pair, we apply a single-point crossover at a randomly chosen crossing point. For each pair, the parent pair survives and the two children will be added to the generation (Fig 4).

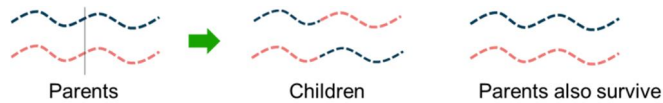

**Fig 4. Sketch of the single-point crossover employed in our algorithm.**

### 3.1.4 Mutation

The mutation in a chromosome is performed for two different factors: (i) for patient IDs and (ii) for gantry statuses. The mutation ratio $r_m$ is preset and its range is typically [0.05, 0.40]. For a generation with $N$ chromosomes, the mutations (i) and (ii) are applied to $r_m N$ randomly-chosen chromosomes.
(i) Mutation in patient IDs: We randomly pick up a cell in a chromosome and put a random patient ID. Then the patient IDs in neighbor cells are also changed in accordance with the consecutiveness of the IDs during the prefixed period of the status.
(ii) Mutation in gantry statuses: We randomly pick up a cell in a chromosome and put a random gantry status. Then the gantry statuses in neighbor cells are also changed in accordance with the prefixed period of the status.

### 3.1.5 Repair Operation

It is usual in evolutionary optimization that repairs of chromosomes are conducted after the crossover and mutation so that otherwise highly-deviated chromosomes are modified to meet the prerequisite condition [18,19]. In this step, for a generation consisting of *N* chromosomes, $r_r N$ chromosomes are randomly chosen where $r_r$ is the repair ratio. Then, their entire time slots are modified to increase the fitness by using the identical conditions that govern the fitness function (as shown in section 3.1.2).

## 3.2 Quantum-Inspired Algorithm

The quantum-inspired algorithm employed here uses a certain number of quantum chromosomes for each generation although it is often discussed that only one or a few are enough [13,14]. The quantum-digit (qudit) representation is employed instead of the common quantum-bit (qubit) representation, which does not affect the algorithmic flow governance. Extra strategies like the inter-group migration strategy [13] and the pair-swap strategy [14] are not applied here. Each step of the algorithm follows its classical counterpart shown in section 3.1 as long as quantum nature allows.

### 3.2.1 Chromosome Design

The quantum-inspired chromosome in our design is depicted in a similar manner as classical one as shown in Fig 5. Each cell for the *g*-th gantry and the *t*-th time slot possesses the two quantum states: (i) A superposition of patient IDs

$$|p(g,t)\rangle = \sum_{j=0}^{n_p-1} \alpha_j(g,t)|j\rangle \qquad (1)$$

and (ii) A superposition of gantry statuses

$$|s(g,t)\rangle = \sum_{k=\text{G\_IDL}}^{\text{G\_PD}} \beta_k(g,t)|k\rangle \qquad (2)$$

where the complex amplitudes $\alpha_j$ and $\beta_k$ should satisfy $\sum_j |\alpha_j|^2 = 1$ and $\sum_k |\beta_k|^2 = 1$. For simplicity of implementation, hereafter we assume that $\alpha_j, \beta_k \in \mathbf{R}$.

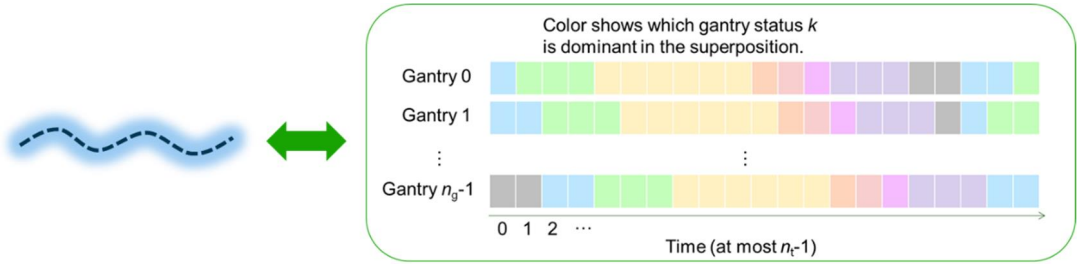

**Fig 5. A quantum chromosome representing the daily schedule.**

### 3.2.2 Selection Rule and the Fitness

The fitness value is computed in the same manner as classical case using Table 2. The only difference is how to determine the patient ID *p* and the gantry status *s* of a cell for each $(g, t)$. We employ a non-demolition projection measurement: we simulate a single-shot projection measurement for equations (1) and (2) and obtain the measurement results *p* and *s*. Nonetheless, quantum states (1) and (2) are kept intact by the simulated projection and reused for the next operation. To simulate a single-shot projection, we use a uniform random number *u* in [0,1]. For the state (1), the result is $j_u$ if the sum of $|\alpha_j|^2$ just exceeds or equals *u* at $j = j_u$ (the sum starts from *j*=0). Similarly, for the state (2), the result is $k_u$ if the sum of $|\beta_k|^2$ just exceeds or equals *u* at $k = k_u$ (the sum starts from *k*=G_IDL).

### 3.2.3 Crossover

The same crossover strategy as the classical counterpart is employed by ignoring the no-cloning theorem.

### 3.2.4 Mutation
The mutation strategy same as the classical counterpart is employed, for which demolition projections are used. For each of the $r_m N$ chromosomes chosen for mutation, we randomly pick up a cell [namely, some $(g, t)$] and the two quantum states written as equations (1) and (2) are projected to random integer states.

### 3.2.5 Repair Operation
In a similar manner to the classical case introduced in section 3.1.5, $r_r N$ chromosomes are randomly chosen. They are individually evaluated using non-demolition measurements and the desirable gantry status and patient ID are determined for each time slot on the basis of the scores in Table 2. Then, the amplitudes of the desirable status and ID are enhanced by the factor of ten or until they reach $\sqrt{1/4}$ at minimum. The other amplitudes are reduced in order to meet the normalization condition of the quantum states of the time slot [see equations (1) and (2)]. This process is a bypassed simulation of a high-dimensional unitary rotation for amplitude amplification and can be regarded as an extension of the single-qubit unitary rotation found in the conventional quantum genetic algorithms.

# 4. Numerical Results
## 4.1 Medium-Size Problem Instance
First, a problem instance with 12 patients is considered. The preset parameter values in our numerical experiments are listed in Table 3. Here, $N_{\text{ini}}$ is the population size of the initial generation and $G_{\text{max}}$ is the maximum number of generations. This parameter set was a test case without prior assessments. Results by a parameter sweep will be shown in section 4.1.1.

**Table 3. The preset values in numerical experiments for the medium-size problem instance.**

| Common | | | | Classical | | Quantum Inspired | |
|---|---|---|---|---|---|---|---|
| $n_g$ | 3 | $r_s$ | 0.83 | $N_{\text{max}}$ | 150 | $N_{\text{max}}$ | 50 |
| $n_p$ | 12 | $r_c$ | 0.27 | | | | |
| $n_t$ | 108 | $r_m$ | 0.37 | | | | |
| $N_{\text{ini}}$ | 10 | $r_r$ | 0.85 | | | | |
| $G_{\text{max}}$ | 200 | | | | | | |

The computation was performed with a single CPU thread on a CPU node of node group A of the Genkai supercomputer of Kyushu University [CPU: 2 × Intel Xeon Platinum 8490H (60 cores, 1.90~3.50 GHz), Memory: 512GiB, Software: Rocky Linux 8, GCC 8.5.0 (C++ language)] using double-precision arithmetics. The running times are shown in Table 4. Note that multi-thread computing was used for a larger-size problem as we will see in section 4.2.

**Table 4. Elapsed time (real CPU time, single thread).**

| Classical | 8.29 [s] |
|---|---|
| Quantum-Inspired | 19.7 [s] |

The fitness value and the number of chromosomes are plotted as functions of generation for our classical algorithm in Fig 6, and for our quantum-inspired algorithm in Fig 7.

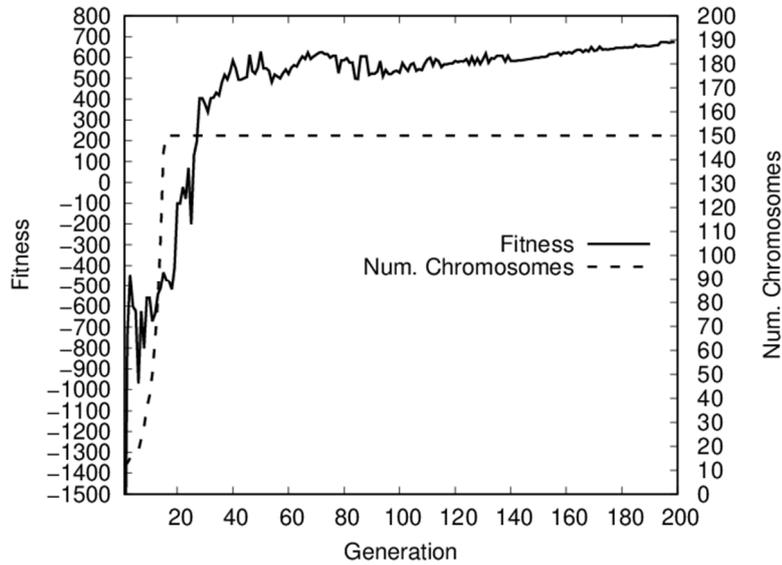

**Fig 6. Fitness of the best chromosome as a function of generation, for our classical genetic algorithm under the setting in Table 3.** The number of chromosomes (namely, the population size) is also plotted as a function of generation.

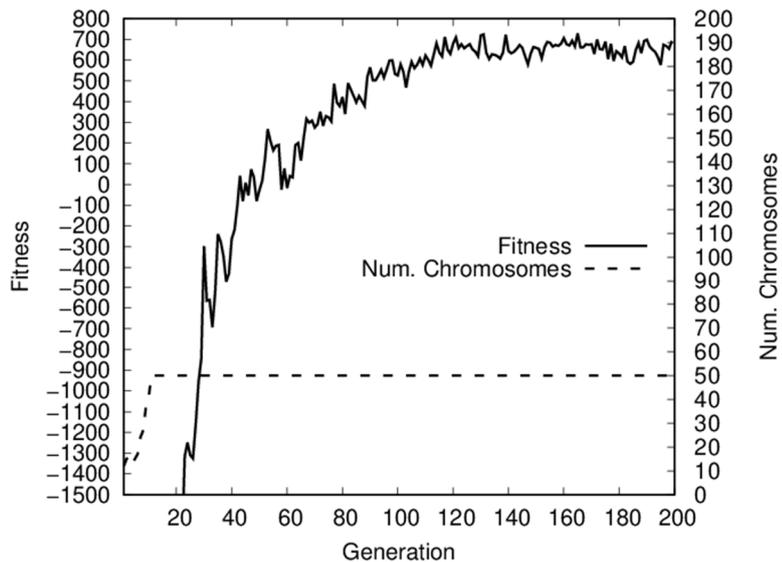

**Fig 7. Fitness of the best chromosome as a function of generation, for our quantum-inspired genetic algorithm under the setting in Table 3.** The number of chromosomes is also plotted as a function of generation.

It has been found that the achieved fitness values were comparable to each other although the quantum-inspired one scored higher during the ripple near generation 190. The highest values were 674 for the classical one and 728 for the quantum-inspired one; both exhibited the convergence to approximately 670. The ripples in the fitness curves were caused by probabilistic mutation affecting even very high-fitness chromosomes.

### 4.1.1 Parameter Sweep

It is common in the research area of genetic algorithms to investigate the effect of parameter values on the achievable fitness values [9]. We performed a parameter sweep over $r_s$, $r_c$, $r_m$, and $r_r$ for our classical and quantum-inspired genetic algorithms within the range of $\pm 0.04$ for each parameter. Fig 8 shows the results as functions of $r_c$ (each data point corresponds to a particular combination of parameter values).

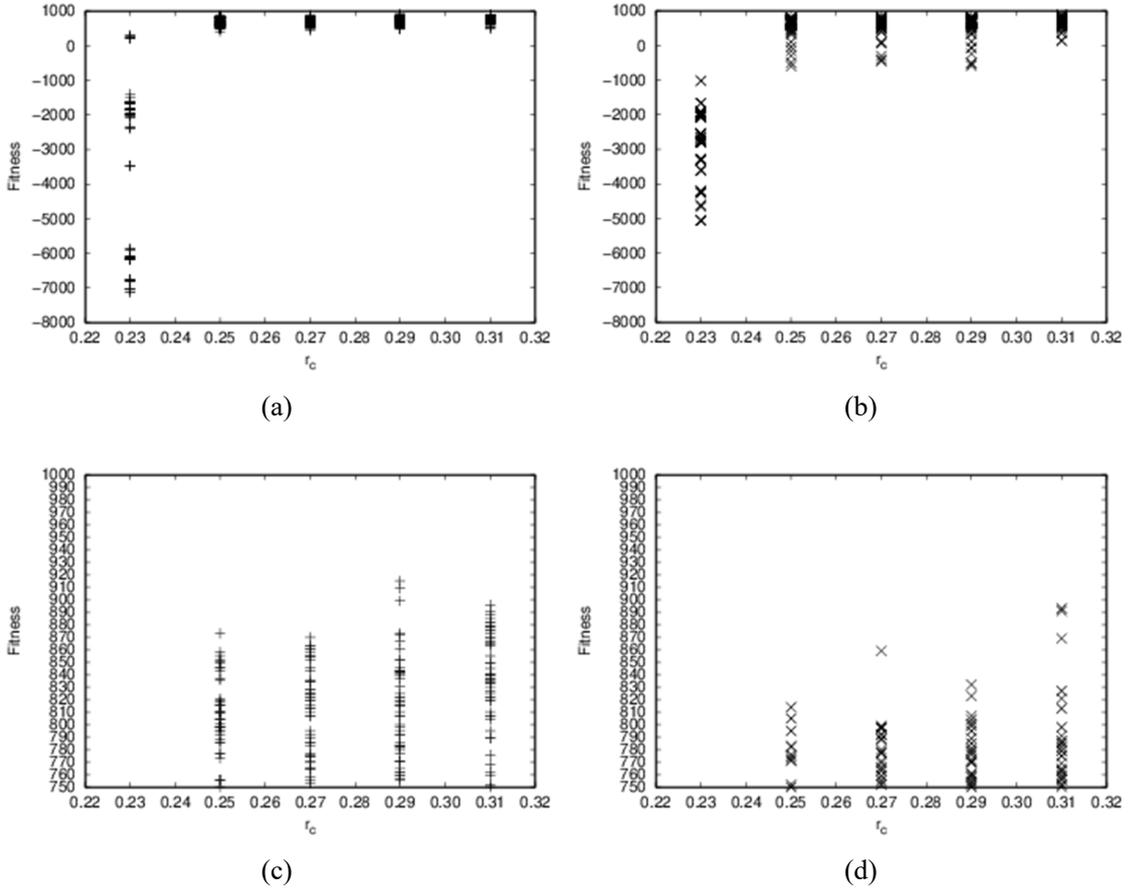

**Fig 8. Plots of fitness data obtained by the parameter sweep for the medium-size problem, shown as functions of $r_c$.** (a) Results for the classical genetic algorithm. (b) Results for the quantum-inspired one. (c) Cutout of (a) for a high-fitness region. (d) Cutout of (b) for a high-fitness region.

The classical one exhibited high fitness values more constantly than the quantum-inspired one. The best achieved data were comparable; fitness value 915 for the classical one (with $r_s$, $r_c$, $r_m$, $r_r$ = 0.85, 0.29, 0.39, 0.89, respectively) and 893 for the quantum-inspired one (with $r_s$, $r_c$, $r_m$, $r_r$ = 0.87, 0.31, 0.39, 0.87, respectively). It should be noted that the quantum-inspired one was run with a smaller population size limit, 50 chromosomes, in contrast to 150 for the classical one.

The statistics of fitness and run time for the data points are shown in Table 5. Here, the data for $r_c$ = 0.23 were regarded as outlier data and excluded because their separation from other data was obvious as depicted in subfigures (a), (b) of Fig 8. There were 444 remaining data for the classical case and 500 for the quantum-inspired case. In addition, the statistics for the top 10 high-fitness data points for each case are shown in Table 6.

**Table 5. Statistics of fitness and run time during the parameter sweep for the medium-size problem.** Each run used a single thread. The number of data is 444 for the classical case and 500 for the quantum-inspired (QI) case (outlier data were excluded).

|  | Average | Max | Min | Standard deviation |
|---|---|---|---|---|
| **Fitness, Classical [a.u.]** | $7.15 \times 10^2$ | 915 | 411 | $1.09 \times 10^2$ |
| **Fitness, QI [a.u.]** | $6.47 \times 10^2$ | 893 | -603 | $2.02 \times 10^2$ |
| **Run time, Classical [s]** | 10.8 | 19.6 | 5.90 | 3.07 |
| **Run time, QI [s]** | 21.1 | 30.0 | 13.7 | 3.47 |

**Table 6. Statistics of top 10 high-fitness data points (for each algorithm) for the medium-size problem.**

|  | Average | Max | Min | Standard deviation |
|---|---|---|---|---|
| **Fitness, Classical [a.u.]** | $8.93 \times 10^2$ | 915 | 879 | $1.15 \times 10^1$ |
| **Fitness, QI [a.u.]** | $8.44 \times 10^2$ | 893 | 813 | $2.95 \times 10^1$ |
| **Run time, Classical [s]** | 16.3 | 18.3 | 13.3 | 1.48 |
| **Run time, QI [s]** | 24.2 | 30.0 | 17.8 | 4.02 |

## 4.2 Large-Size Problem Instance

Second, a problem instance with 72 patients is considered, which is realistic considering the present daily operations in radio-therapy facilities. The preset parameter values are shown in Table 7. This is, again, a test case and the assessment by parameter sweep will be shown in section 4.2.1.

**Table 7. The preset values in numerical experiments for the large-size problem instance.**

| Common | | | | Classical | | Quantum Inspired | |
|---|---|---|---|---|---|---|---|
| $n_g$ | 3 | $r_s$ | 0.83 | $N_{ini}$ | 40 | $N_{ini}$ | 10 |
| $n_p$ | 72 | $r_c$ | 0.37 | $N_{max}$ | 250 | $N_{max}$ | 70 |
| $n_t$ | 650 | $r_m$ | 0.37 | | | | |
| $G_{max}$ | 200 | $r_r$ | 0.85 | | | | |

The computation was performed with 30 threads (using OpenMP) on a CPU node of node group A of the Genkai supercomputer using double-precision arithmetics (see section 4.1 for node specification). The running times are shown in Table 8.

**Table 8. Elapsed time (real time using 30 OpenMP threads).**

| Classical | 261 [s] |
|---|---|
| Quantum-Inspired | 1478 [s] |

The fitness value and the number of chromosomes are plotted as functions of generation for our classical algorithm in Fig 9, and for our quantum-inspired algorithm in Fig 10. Let us mention again that the ripples of fitness curves are owing to probabilistic mutations affecting even high-fitness chromosomes.

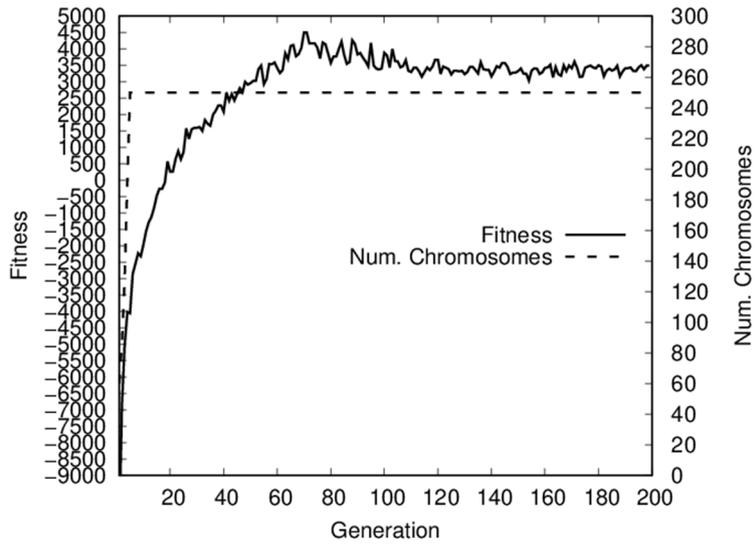

**Fig 9. Fitness of the best chromosome as a function of generation, for our classical genetic algorithm under the setting in Table 7.** The number of chromosomes is also plotted as a function of generation.

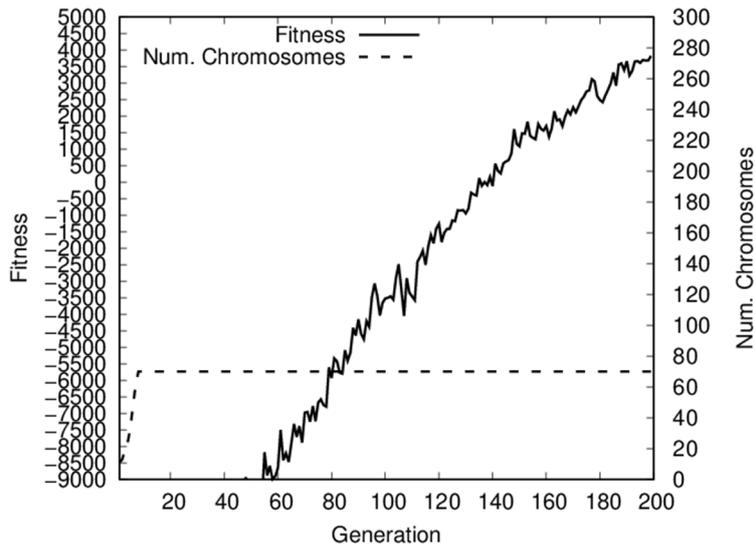

**Fig 10. Fitness of the best chromosome as a function of generation, for our quantum-inspired genetic algorithm under the setting in Table 7.** The number of chromosomes is also plotted as a function of generation.

It has been found that both algorithms achieved the high fitness value in the range from 3500 to 4500 and the classical one exhibited a much faster convergence. The highest fitness values were 4499 for the classical one and 3818 for the quantum-inspired one. It should be, however, noted that the number of chromosomes went up to 250 for the classical one while only 70 for the quantum-inspired one.

## 4.2.1 Parameter Sweep

We performed a parameter sweep over $r_c$ and $r_m$ for our algorithms within the range of $\pm 0.05$ for $r_c$ and $0.17 \leq r_m \leq 0.62$, with $r_s$ and $r_r$ fixed as they are in Table 7. Fig 11 shows the results as functions of $r_c$ (each data point corresponds to a particular combination of parameter values). The results illustrated in Fig 11 shows that the classical genetic algorithm achieved high fitness values in a stable manner while the quantum-inspired one rather unstable. It however also shows that the quantum-inspired one achieved higher fitness values as in subfigures (c) and (d). The classical one achieved its highest value 3913 (with $r_s$, $r_c$, $r_m$, $r_r$ = 0.83, 0.42, 0.47, 0.85, respectively); in contrast, the quantum-inspired one achieved 4836 (with $r_s$, $r_c$, $r_m$, $r_r$ = 0.83, 0.42, 0.22, 0.85, respectively). The smaller population size for the quantum-inspired one was also its advantage (it was 70 in contrast to the classical one's 250).

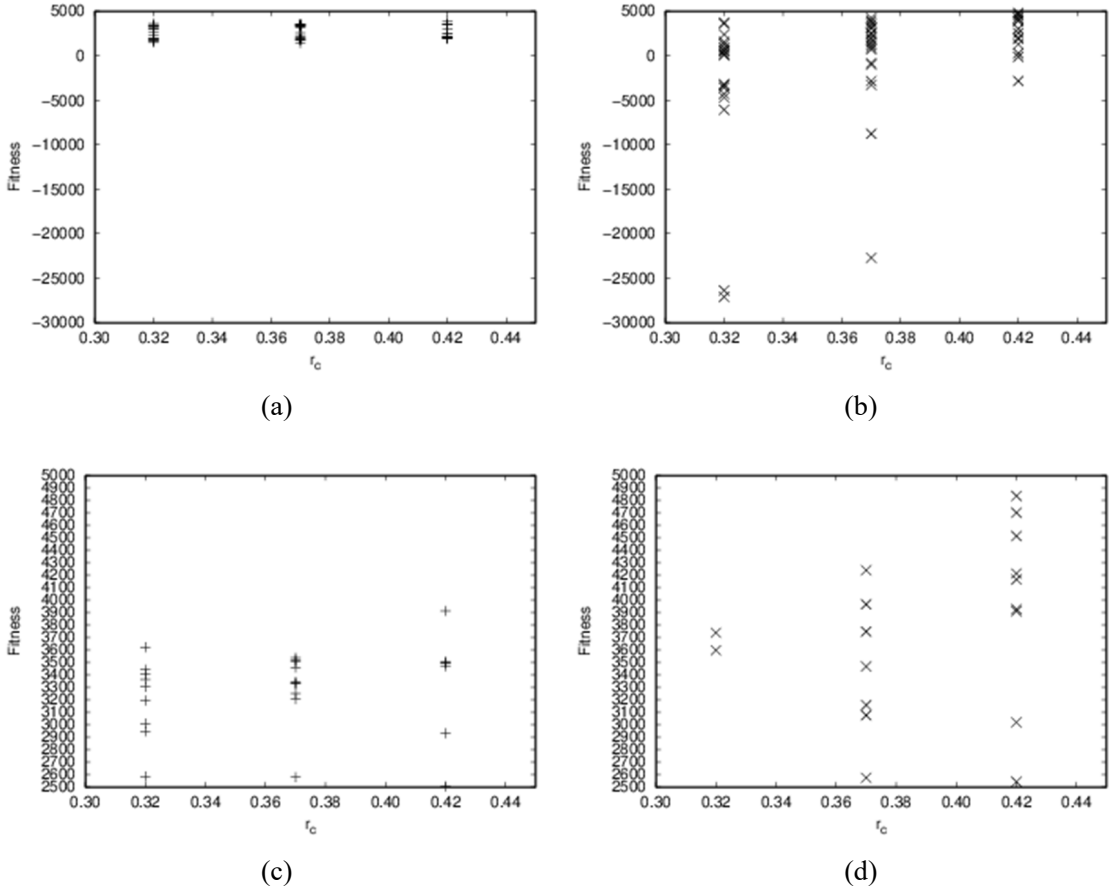

**Fig 11. Plots of fitness data obtained by the parameter sweep for the large-size problem, shown as functions of $r_c$.** (a) Results for the classical genetic algorithm. (b) Results for the quantum-inspired one. (c) Cutout of (a) for a high-fitness region. (d) Cutout of (b) for a high-fitness region.

The statistics of fitness and run time for the data points are shown in Table 9. There were 54 data for the classical case and the same for the quantum-inspired case. In addition, the statistics of top ten high-fitness data points for each case are shown in Table 10.

**Table 9. Statistics of fitness and run time during the parameter sweep for the large-size problem.** Each run used 30 OpenMP threads. The number of data is 54 for both classical and quantum-inspired (QI) cases.

|  | Average | Max | Min | Standard deviation |
|---|---|---|---|---|
| **Fitness, Classical [a.u.]** | $2.59 \times 10^3$ | 3913 | 1422 | $7.26 \times 10^2$ |
| **Fitness, QI [a.u.]** | $-5.48 \times 10^2$ | 4836 | -27106 | $6.73 \times 10^3$ |
| **Run time, Classical [s]** | $6.13 \times 10^2$ | $3.57 \times 10^3$ | $1.04 \times 10^2$ | $6.21 \times 10^2$ |
| **Run time, QI [s]** | $1.41 \times 10^3$ | $2.71 \times 10^3$ | $1.08 \times 10^1$ | $7.83 \times 10^2$ |

**Table 10. Statistics of top 10 high-fitness data points (for each algorithm) for the large-size problem.**

|  | Average | Max | Min | Standard deviation |
|---|---|---|---|---|
| **Fitness, Classical [a.u.]** | $3.55 \times 10^3$ | 3913 | 3459 | $1.28 \times 10^2$ |
| **Fitness, QI [a.u.]** | $4.22 \times 10^3$ | 4836 | 3747 | $3.42 \times 10^2$ |
| **Run time, Classical [s]** | $9.44 \times 10^2$ | $1.77 \times 10^3$ | $2.95 \times 10^2$ | $5.65 \times 10^2$ |
| **Run time, QI [s]** | $2.15 \times 10^3$ | $2.71 \times 10^3$ | $1.19 \times 10^3$ | $6.27 \times 10^2$ |

# 5. Discussion

We have introduced a tailored quantum-inspired genetic algorithm for the patient-scheduling problem in proton therapy. It should be noted that our algorithm is applicable to a clinical scheduling problem with a similar problem structure, especially that for photon therapy by modifying the durations in Table 1 in accordance with its clinical practice.

With a numerical exploratory investigation, it has been shown that a quantum-inspired genetic algorithm achieves comparable fitness values in comparison to its classical counterpart for a medium-size therapy scheduling problem under a similar order of CPU time consumption using a smaller population size. This result indicates, albeit empirically, that neither the group migration strategy nor the pair swap strategy might be necessary for handling scheduling problems apart from the convergence speed dispute. The convergence time deviation within a few minutes is not critical to the problem of our interest considering the clinical administration time scale. Patient scheduling in radiotherapy centers is normally finalized ahead of the clinical day.

A similar tendency has been observed for a large-size problem corresponding to a real clinical circumstance. Furthermore, the quantum-inspired one outperformed to some extent in the fitness values reached by a parameter sweep. A drawback is the CPU time consumption as the quantum-inspired one took approximately 23.5 minutes on average for a single run. This is an unexpected result in light of an established standpoint in this field that quantum-inspired genetic algorithms should run with smaller computational costs in comparison to classical counterparts in optimization problems (see [20,21] for the reported superiority of quantum-inspired ones). It was indeed reported that a few advanced classical methods sometimes outperformed quantum-inspired ones [15], but the present classical counterpart had a very common algorithmic structure. The reason for the worse time performance of our quantum-inspired genetic algorithm is that we could not afford to handle many long quantum chromosomes in short time even though we used a node of a supercomputer. This is certainly a limit of classical computers when trying to emulate quantum nature for large-size problem instances.

A future direction may involve the use of a real quantum computer to handle the scheduling problem with an evolutionary algorithm. It seems, however, not a near future that a real quantum computer will have enough resource. The required number of qubits is easily estimated. For each chromosome, we need
$$n_t n_g (\lceil \log_2 n_p \rceil + \lceil \log_2 n_s \rceil)$$
qubits where $n_s$ is the number of gantry statuses. For a generation with population size *N*, we need
$$N n_t n_g (\lceil \log_2 n_p \rceil + \lceil \log_2 n_s \rceil)$$
qubits. Using the values found in Tables 1 and 7, it is found that approximately $1.4 \times 10^6$ qubits are required. This is beyond the near-term quantum technology [22,23] and it is hoped to be realized within several decades from now. A technique to use a real quantum computer for evolutionary computing [24-26] will then be useful.

Another point we should state is that the stability of the standard classical genetic algorithm has been reconfirmed through this study. It exhibited a steady production of relatively high-fitness schedules, which is a preferable feature in case one has no access to high-performance computers.

# 6. Conclusion

We investigated the applicability of a quantum-inspired genetic algorithm for the schedule optimization problem for proton therapy assuming a setup in a radiation oncology center, in comparison to a classical counterpart. Our quantum-inspired algorithm was found to be a usable option in the presently available computer resources; it generated high-fitness schedules within practical time using a node of a supercomputer although its merit was insignificant in comparison to the classical counterpart currently. Its run time is expected to be reduced when it is updated to use real quantum resources along with the paradigm shift targeting the quantum era.

# Acknowledgements

AST was supported by KAKENHI grant from JSPS, Japan (No. 18K11344).